\documentclass[sn-basic]{sn-jnl}
\usepackage[T2A]{fontenc}
\usepackage[utf8]{inputenc}
\usepackage{caption}
\usepackage{array}
\usepackage[super]{nth}
\usepackage{hyperref}
\usepackage{graphicx,subcaption}	
\usepackage{xcolor}
\usepackage{soul} 
\usepackage{arabtex}
\usepackage{utf8}
\setcode{utf8}

\soulregister\cite7
\soulregister\citep7
\soulregister\cref7

\usepackage[capitalise, noabbrev]{cleveref}
\crefname{subsection}{Subsection}{Subsections}

\theoremstyle{thmstyleone}%
\theoremstyle{thmstyletwo}%

\theoremstyle{thmstylethree}%

\begin{document}
\title[
Mismatching-Aware Unsupervised Translation Quality Estimation...
]{
Mismatching-Aware Unsupervised Translation Quality Estimation For Low-Resource Languages
}


\author[1]{\fnm{Fatemeh} \sur{Azadi}}\email{ft.azadi@ut.ac.ir}

\author*[1]{\fnm{Heshaam} \sur{Faili}}\email{hfaili@ut.ac.ir}

\author[1]{\fnm{Mohammad~Javad}~\sur{Dousti}}\email{mjdousti@ut.ac.ir}

\affil*[1]{ \orgdiv{School of Electrical and Computer Engineering}, \orgdiv{College of Engineering}, \orgname{University of Tehran}, \orgaddress{\city{Tehran}, \country{Iran}}}


\abstract{Translation Quality Estimation (QE) is the task of predicting the quality of machine translation (MT) output without any reference. This task has gained increasing attention as an important component in the practical applications of MT. In this paper, we first propose XLMRScore, which is a cross-lingual counterpart of BERTScore computed via the XLM-RoBERTa (XLMR) model. This metric can be used as a simple unsupervised QE method, 
nevertheless facing two issues: firstly, the untranslated tokens leading to unexpectedly high translation scores, and secondly, the issue of mismatching errors between source and hypothesis tokens when applying the greedy matching in XLMRScore. To mitigate these issues, we suggest replacing untranslated words with the unknown token and the cross-lingual alignment of the pre-trained model to represent aligned words closer to each other, respectively.
We evaluate the proposed method on four low-resource language pairs of the WMT21 QE shared task, as well as a new English$\rightarrow$Persian (En-Fa) test dataset introduced in this paper.
Experiments show that our method could get comparable results with the supervised baseline for two zero-shot scenarios, i.e., with less than 0.01 difference in Pearson correlation, while outperforming unsupervised rivals in all the low-resource language pairs for above 8\%, on average. }




\maketitle

\section{Introduction}\label{sec1}
Translation quality estimation (QE) is defined as the task of determining the quality of machine translation (MT) models without the need for gold reference translations \citep{specia2009}. The reference-based evaluation metrics commonly used for automatic evaluation of MT, such as BLEU \citep{papineni2002}, METEOR \citep{banerjee2005} or TER \citep{snover2006}, score the MT outputs in comparison to a gold reference. However, references are not always available, especially in real-time scenarios where we want to inform the user about the reliability of MT outputs or to increase the efficiency in automatic post-editing \citep{do2021review} (e.g., by estimating the effort needed to post-edit each sentence). Thus, reference-free evaluation metrics (i.e., QE) have become a significant area of research in the field of MT.
QE can be performed at various levels, including word, sentence, or document level \citep{kepler-etal-2019-unbabels}. The sentence level QE, which is the focus of this paper, is defined as predicting either of the two objectives of \textit{Direct Assessment} (DA), which is the perceived quality score of translated sentences or post-editing effort, which is the number of edits a translator needs to achieve a perfect translation measured by the \textit{Human-mediated Translation Edit Rate} (HTER) metric \citep{snover2006}.

State-of-the-art (SOTA) QE models often need several thousand pairs of sentences (more than 7K sentence pairs in standard datasets) with DA or HTER quality labels for training \citep{specia2021findings}. However, such annotated data is not easily obtainable for most languages, especially in low-resource scenarios. Thus, unsupervised approaches without the need for any QE supervision would be beneficial and attractive.

BERTScore \citep{zhang2020bertscore} is a reference-based evaluation metric, which is mostly suitable for monolingual scenarios, i.e., the scenarios in which the hypothesis and the reference are in the same language. Conversely, QE is a cross-lingual task where we evaluate the MT output with respect to the input sentence, which is in another language. However, BERTScore can be applied to QE using cross-lingual models, i.e., models that represent mappings between words in different languages. 
The idea of using the cross-lingual counterpart of BERTScore as a fully unsupervised sentence-level QE metric was firstly adopted by \cite{zhou2020zero}, where they used the multilingual BERT (MBERT) model \citep{devlin2019bert} to compute BERTScore between source and translated sentences.
Following \cite{zhou2020zero}, in this paper we use the XLMR-Base model instead of MBERT, because it was shown to provide much better representations for low-resource languages~\citep{conneau2020unsupervised}, while noticing that computing BERTScore with these multilingual models and applying it directly to the QE task will cause certain issues. We target these issues in this paper and propose two strategies to mitigate them.


The first issue is that many mismatching errors arise when performing greedy matching between source and translated candidate tokens with pairwise word similarities in BERTScore. This problem came to light and was initially termed the \textit{mismatching issue} by \cite{zhou2020zero}. This issue occurs due to the fact that the utilized models, i.e., MBERT or XLMR, are multilingual rather than cross-lingual.
In other words, they are just pre-trained jointly on multiple languages over a set of monolingual data for each language and without any cross-lingual supervision of mappings between them. Although these models have shown impressive cross-lingual ability in various tasks \citep{k-etal-2020, wu-dredze-2019-beto, conneau2020unsupervised}, they could not perfectly represent mappings between words in different languages and thus the mismatching issue occurs.
%
To alleviate this issue, we propose to incorporate the cross-lingual supervision of word alignments and make the pre-trained model more cross-lingually aligned.
The resultant model becomes more capable of providing similar representations for aligned words in different languages. We called this approach \textit{cross-lingual alignment of the pre-trained model}.
For this purpose, we fine-tune the base model, i.e., the XLMR-Base in our case, on word alignments derived from a statistical aligner like GIZA\texttt{++} on a parallel corpus. Our fine-tuning approach is a contrastive learning approach, as in \cite{wudredze2020}, where the base model was optimized to bring the representation of aligned words closer together while moving the unaligned ones further apart. However, our approach to fine-tuning differs from \cite{wudredze2020} in selecting negative examples and considering all subwords for alignment, according to our application.

The second issue, which we encounter while using BERTScore for QE, is untranslated words.
Because these words have very close representations to their corresponding source words, the model could not recognize them and would assign a high similarity score to them in the greedy matching. To mitigate this issue, we create a vocabulary from a large monolingual corpus of the target language, assuming the untranslated words in the hypothesis as the words that are not in this vocabulary and replacing them with the unknown token, i.e., $\langle unk \rangle$.


To conduct our experiments, we use test sets from the WMT21 shared task on QE \citep{fomicheva2022mlqe} for low-resource language pairs of Sinhala$\rightarrow$English, Nepalese$\rightarrow$English, Pashto$\rightarrow$English, and Khmer$\rightarrow$English\footnote{Arrows indicate translation directions.}. We also create a new test set for the English$\rightarrow$Persian (En-Fa) language pair, which is annotated with the post-editing metric of HTER and evaluate our proposed QE metric on it.
This test set is used for the WMT23 QE shared task\footnote{\url{https://wmt-qe-task.github.io/}} and is therefore publicly available. Our implementations are also publicly available\footnote{\url{https://github.com/fatemeh-azadi/Unsupervised-QE}}. Furthermore, as some of these languages are somehow close to each other, like Persian and Pashto, we try to examine if using the word alignment knowledge of similar languages could help. To do this, we fine-tune the XLMR model on word alignments from all these languages as well as the word alignments derived from an English-Hindi corpus simultaneously. Hindi is similar to Persian, Pashto, Nepalese, and Sinhala languages and additionally has more clean parallel resources available, which can be beneficial for attaining high-quality word alignments.

 
The main contributions of our work can be summarized as follows. 

\begin{itemize}
\item Proposed methods to tackle issues of untranslated words and mismatching errors while using the cross-lingual counterpart of BERTScore as an unsupervised QE method for low-resource languages.
\item Studied the impact of using closely related languages while fine-tuning the pre-trained models to have more cross-lingually aligned contextual word embeddings.
\item Provided the first test set with HTER labels for the English$\rightarrow$Persian language pair, suitable for the sentence-level post-editing effort task as in the WMT QE shared task.
\end{itemize} 

The rest of this paper is organized as follows. In the next section, we briefly discuss the related work. In \cref{sec3}, we will describe our base method while arguing issues and suggesting our solutions. Next, we introduce the datasets that we used in this paper, including the new English$\rightarrow$Persian test set in \cref{sec4}, and describe our experimental setup in \cref{sec5}. Afterwards, we present our main results in \cref{sec6}, and provide further analysis and discussions in \cref{sec7}. Finally, we conclude the paper while providing an outlook for future work in the last section.

\section{Related Work}\label{sec2}

QE refers to the task of predicting MT quality without any translation reference. Traditional approaches for QE use conventional machine learning models while designing various manual features. Such features are extracted from the MT system, the source and translated sentence, and external resources such as monolingual or parallel corpora \citep{specia2009, specia2013}. SOTA approaches use neural methods. \cite{kim2017predictor} proposed the predictor-estimator model, which is still used by the most recent SOTA methods \citep{specia2021findings}. In \cite{kim2017predictor}, the predictor was an encoder-decoder RNN, which randomly selects and masks a word in a target sentence from parallel data and learns to predict it. The estimator was also an RNN which uses representations generated by the predictor to learn QE scores. Similar approaches were also proposed by \cite{ive2018} as a single end-to-end model and \cite{wang2018}, which first used a transformer-based model to extract features and then fed these features with manually designed features into a BiLSTM to train for QE.

Advances in large-scale multilingual pre-trained models have greatly improved the performance of recent QE methods. \cite{kepler-etal-2019-unbabels} used the predictor-estimator while replacing the predictor with the BERT or XLM \citep{conneau2019cross} model. \cite{ranasinghe2020} fine-tuned a single or two separate XLMR models for source and target sentences on QE scores. \cite{lee2020} generated a huge artificial QE dataset based on a parallel corpus, pre-trained the XLMR model on it, and fine-tuned it using the QE train dataset. There are also many other supervised methods proposed using the pre-trained language models \citep{kim2019, moura2020, chen2021, wang2021}. Despite achieving strong performance, supervised approaches need QE-labeled data for training, which is not easily obtainable for most languages. Thus zero-shot or unsupervised QE has become an attractive research area.

Several studies have attempted to develop unsupervised QE methods in recent years; however, their performance is inferior to the SOTA supervised approaches. \cite{etchegoyhen2018} used lexical translation and statistical language model probabilities while averaging them to get the final QE scores. \cite{fomicheva2020unsupervised} used various glass-box features extracted from NMT models to propose an unsupervised QE model. \cite{zhou2020zero} proposed an enhanced version of BERTScore, which uses word alignments from a statistical word alignment model trained on a parallel corpus to define a penalty function over the similarity of unaligned tokens in the greedy matching of BERTScore. They also defined a generation score as the perplexity obtained for each target side token and combined it with the BERTScore value to get the unsupervised QE score. While their work is the most similar to ours in using the cross-lingual counterpart of BERTScore, unlike them, we propose to use more cross-lingually aligned word embeddings instead of penalizing the similarity of unaligned tokens to alleviate the mismatching issue. Furthermore, We reveal the untranslated words issue while providing a solution to mitigate it. 

\cite{tuan2021} proposed two methods to generate synthetic data for QE. Using a parallel corpus, in the first method, they translated source sentences with an NMT system, and in the second, they injected errors into target sentences with a masked language model. Afterwards, by considering these outputs as candidate translations and target sentences as pseudo post-edits, they computed HTER tags and trained a predictor-estimator model \citep{kim2017predictor} over this synthetic data instead of human-labeled data. However, their approach is quite different from ours, as our approach is about providing an unsupervised QE metric rather than generating synthetic data. Moreover, their approach is only suitable for the HTER task, as their methods cannot generate DA labels. Unlike \cite{tuan2021}, our proposed method does not rely on HTER tags.
%
Moreover, their work lacks analysis and testing on real low-resource language pairs, while we try to experiment with and analyze real low-resource scenarios in this paper. SMOB-ECEIIT is a fully unsupervised method that participated in WMT21~\citep{specia2021findings}.
It proposed two methods to compute the distance between source and translated sentences and combined these distances linearly to get the final QE scores. The first method calculated the Sinkhorn distance \citep{cuturi2013}, and the second one assumed that each sentence is represented by the distances between its own words and used the Wasserstein distance between the Persistent Homology \citep{Edelsbrunner2012} method outputs to compare the distance matrices of two sentences. We compare our results with this method in \cref{sec6.2}, which shows that we could surpass them on all the targeted low-resource language pairs from WMT21.

Although our unsupervised QE method could not outperform the SOTA supervised models, it is applicable as an effort towards the unsupervised approaches since they are still rather under-explored and need more attention. Also, as discussed in \cref{sec7.3}, the success of the supervised methods on low-resource and zero-shot language pairs, such as the ones in the WMT21 test data that we used in our experiments, is partly due to their shared target side language, i.e., English, with the available training data in the WMT20 and WMT21 QE shared tasks \citep{specia-etal-2020}. Thus, working on unsupervised approaches can be attractive and more appropriate for the target side languages with no training data available.

\section{Proposed Method}\label{sec3}
The base of our unsupervised QE method is the well-known BERTScore metric. Nonetheless, in this context, we refer to it as \textit{XLMRScore} because we opted for the XLMR model over M-BERT due to its superior performance in capturing meaningful representations for low-resource languages. This was shown in \cite{conneau2020unsupervised}, where the XLMR model could significantly outperform the M-BERT model on a variety of cross-lingual benchmarks, particularly for low-resource languages.

In this section, we first describe the base XLMRScore method in detail while discussing the issues we will encounter when using it directly for the QE purpose. Next, we introduce our proposed solutions to replace untranslated words and cross-lingual alignment of the pre-trained model to alleviate the issues of untranslated words and mismatching errors.


\subsection{XLMRScore}\label{subsec3.1}
BERTScore was proposed by \cite{zhang2020bertscore} as a reference-based evaluation metric for any text generation task, including MT. It focused on computing semantic similarity between the tokens of reference and hypothesis, using the contextual word embeddings derived from the pre-trained BERT model. To use it for the QE task with no reference, we could use pre-trained multilingual models like M-BERT as in \cite{zhou2020zero}, and calculate the pairwise similarity of tokens in the source and hypothesis sentences. Then by using greedy matching, each token in the source sentence matched to the closest token in the hypothesis, and vice versa, to compute the recall and precision scores.

In more detail, using the pre-trained multilingual model, each source sentence is mapped to a sequence of vectors $\langle x_1, \dots, x_n \rangle$, while its corresponding hypothesis is mapped to $\langle y_1, \dots, y_m \rangle$. Then the QE scores based on BERTScore can be computed as follows:

\begin{align}
P_{} &= \frac{1}{\vert y \vert} \sum\limits_{y_j \in y} \max\limits_{x_i \in x} x_i^\top y_j,  \nonumber \\
R_{} &= \frac{1}{\vert x \vert} \sum\limits_{x_i \in x} \max\limits_{y_j \in y} x_i^\top {y}_j, \label{eq1}
\end{align}

\noindent where $P$ and $R$ represent precision and recall rates, respectively. In this paper, we use these scores as our base QE metrics. However, we note that the original BERTScore, proposed by \cite{zhang2020bertscore}, is suitable for use in monolingual scenarios, and applying it directly to a cross-lingual setting like QE, as described above, causes two issues:

\begin{enumerate}

    \item \textbf{Untranslated tokens:}
    There may be some untranslated tokens in the MT output that actually belong to the source language. Thus, their representations will be very close to their corresponding source equivalents, leading to a very high quality score, which is undesirable.
    
    \item \textbf{Mismatching errors:}
    As indicated in \cite{zhou2020zero}, many source or hypothesis tokens are not correctly matched to their corresponding token on the other side when using the greedy matching approach described above. This leads to errors in computing the QE scores. These errors, called mismatching errors, are related to the utilized multilingual model, which cannot provide close representations for some of the corresponding tokens and thus cannot match them to each other correctly.

\end{enumerate}

To resolve these issues, we propose these two strategies, which will be explained extensively in the following subsections:
\begin{enumerate}
    \item Replacing untranslated tokens with the unknown token (\cref{subsec3.2}) 
    \item Cross-lingual alignment of the base pre-trained multilingual model (\cref{subsec3.3}) 
\end{enumerate}


We have performed preliminary experiments using both M-BERT and XLMR-Base models to compute the aforementioned QE scores. As expected, M-BERT produced worse results than those generated by XLMR-Base since it provides worse representations for low-resource languages than XLMR-Base. This was previously indicated in \cite{conneau2020unsupervised} and is mostly due to the difference in scale of the training data they used. M-BERT was pre-trained on Wikipedia, while XLMR-Base was pre-trained on a cleaned CommonCrawl corpus, which provides a significantly larger amount of data than Wikipedia, especially for low-resource languages.
%
Thus, we use the XLMR-base model for the rest of our experiments, and hence we call the base method XLMRScore instead of BERTScore. We use the precision rate in \cref{eq1} as the final metric in our reported results for WMT21 because it shows better correlations with the golden QE labels in our experiments.

\subsection{Replacing untranslated tokens}\label{subsec3.2}

Using XLMRScore for QE is based on the assumption that semantically similar tokens in the source and target languages have similar representations in the utilized model. However, the popular multilingual models like XLMR do not have any signal about the language of the tokens.
Thus, untranslated tokens cannot be detected properly and will be matched with their corresponding source token with a high similarity score.
The extreme case is when the translation model cannot translate the source sentence entirely, and the hypothesis is exactly the same as the source sentence. In this case, computing XLMRScore in the way above will give the highest score of 1 as the QE score, which is not appropriate.

To mitigate this issue, we force the model to treat the untranslated tokens as out-of-vocabulary in the target language. For this purpose, we create a vocabulary by considering the words with occurrences of more than a certain threshold in a monolingual corpus of the target language. Next, we replace the out-of-vocabulary words in the translation candidates with the unknown token (i.e., $\langle \texttt{UNK} \rangle$ in XLMR model) and score this new translation candidate instead of the original one. More details about statistics of the utilized monolingual corpora and vocabulary are provided in \cref{sec4.3}.

\subsection{Cross-lingual alignment of the pre-trained model}\label{subsec3.3}

The widely used multilingual models, like M-BERT and XLMR, are just pre-trained multilingually and with no cross-lingual supervision, which means they are only pre-trained on concatenated data of multiple languages without any explicit mappings or alignments between different languages. This is why these models are called multilingual rather than cross-lingual, which refers to the models that can represent mappings between words in different languages. Nevertheless, multilingual models were shown to have surprising effectiveness for cross-lingual applications, like zero-shot cross-lingual transfer when fine-tuned on downstream tasks \citep{k-etal-2020, wu-dredze-2019-beto} or 
obtaining word alignments in a parallel corpora \citep{sabet2020simalign}. This suggests that these models are somewhat cross-lingually aligned. 

\begin{figure}[t]%
\centering

\medskip

\begin{subfigure}[t]{.49\linewidth}
\centering\includegraphics[width=\linewidth]{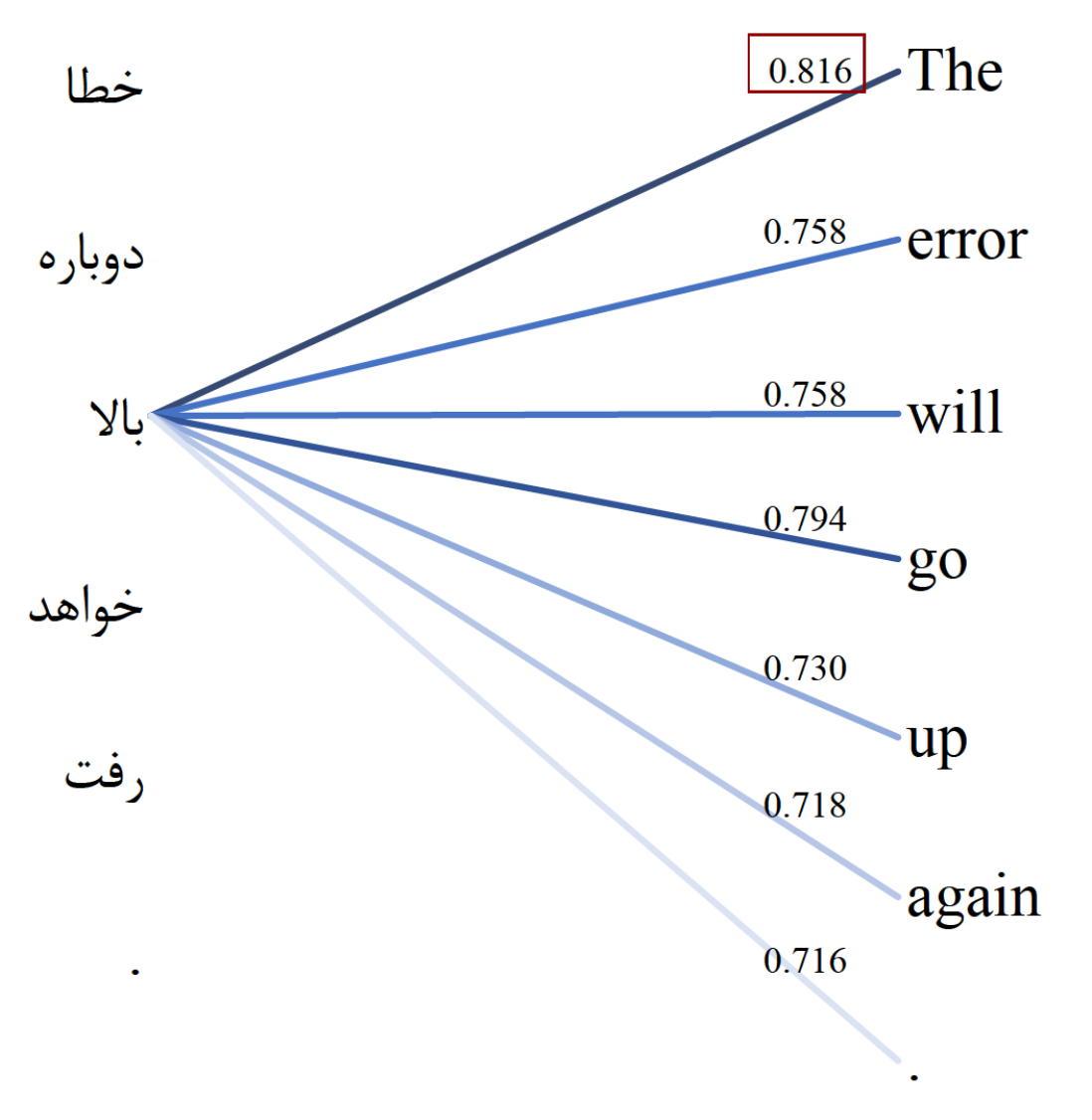}
\caption{Before cross-lingual alignment}
\end{subfigure}
\begin{subfigure}[t]{.49\linewidth}
\centering\includegraphics[width=\linewidth]{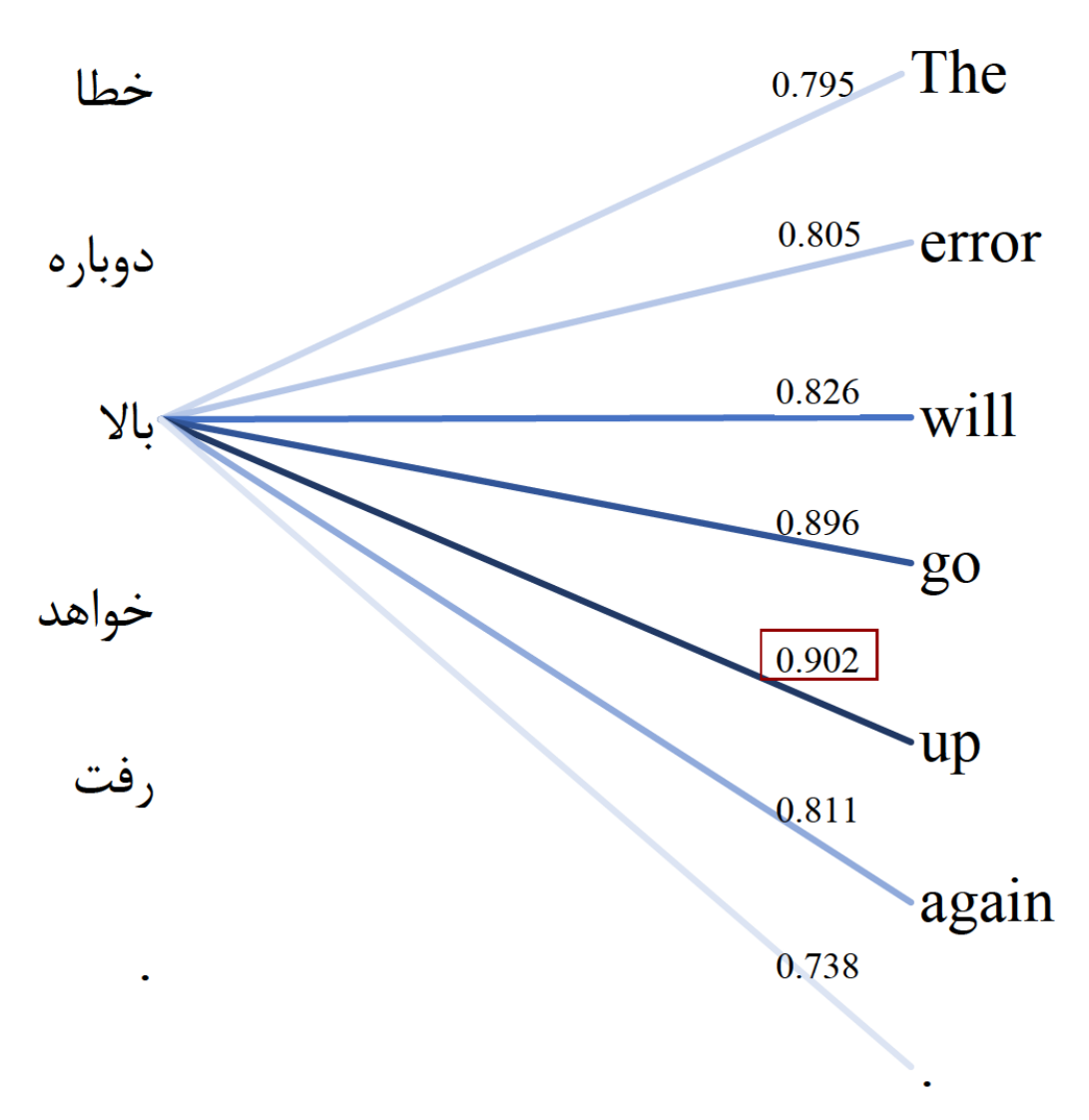}
\caption{After cross-lingual alignment}
\end{subfigure}
\caption{An example of the mismatching issue, where the Persian token ``\<بالا>'' is matched to the English token ``The'' instead of the correct token ``up'' using the base pre-trained model (a). After fine-tuning the model using the cross-lingual alignment strategy, it could correctly match ``\<بالا>'' and ``up'' to each other (b).}
\label{fig1}
\end{figure}

In the context of pre-trained models, a model is ``cross-lingually aligned'' or is in ``cross-lingual alignment'' if it can generate similar vector representations for tokens with the same meaning and context in different languages. This capability has been discussed in \cite{kulshreshtha-etal-2020, wudredze2020} and \cite{cao2019}, while it was targeted for improvement by adding cross-lingual supervisions, and as mentioned in the previous section, it is the assumption behind using XLMRScore for QE. We assume that by greedy matching of tokens, i.e., matching each token with the most similar token in the other side, we match each token with its translation and their similarity shows how good this translation is. However, as shown in \cref{fig1} and indicated by \cite{zhou2020zero}, there are many tokens that are not correctly matched to their corresponding token on the other side when using the greedy matching approach, which are called mismatching errors. These errors will directly affect our QE scores and can make them erroneous.
Mismatching errors occur due to the imperfect cross-lingual alignment of the pre-trained models, which means they could not generate proper similar representations for aligned word pairs within parallel sentences. As stated above, this is because these models were not given any information about word alignments during pre-training.

One way to improve the cross-lingual alignment of pre-trained models is to use explicit cross-lingual objectives, i.e., enforcing similar words from different languages to have similar representations. Such objectives have been used effectively during pre-training \citep{conneau2019cross, huang2019unicoder}, to learn linear mappings \citep{wang2019, aldarmakidiab2019, liu2019}, or in a fine-tuning phase \citep{cao2019, wudredze2020}.

In this work, we follow the contrastive learning approach proposed in \cite{wudredze2020} to fine-tune our base multilingual model. Given the aligned word pairs from a parallel corpus, this approach encourages the model to bring representations of aligned words closer than unaligned ones in a sentence pair. More specifically, we first use GIZA\texttt{++}, which is a statistical word aligner, to derive word-level alignments from a parallel corpus. To improve the precision of these alignments, they are obtained in both the source-to-target and target-to-source directions, and only the ones in the intersection are retained \cite{ochney2003}.

The objective function used for fine-tuning is the same as the contrastive loss proposed in \cite{wudredze2020}.
Assume we have a set of aligned word pairs from GIZA\texttt{++} for each parallel sentence. Considering the $i$th word pair in this set, $(s^{\scriptscriptstyle\parallel}_i, t^{\scriptscriptstyle\parallel}_i)$ shows the corresponding representation vectors from a pre-trained multilingual model.
The contrastive alignment optimizes the relative distance between $s^{\scriptscriptstyle\parallel}_i$ and $t^{\scriptscriptstyle\parallel}_i$, which means enforcing $s^{\scriptscriptstyle\parallel}_i$ to be closer to $t^{\scriptscriptstyle\parallel}_i$ than any other $t^{\scriptscriptstyle\parallel}_j$, $\forall j \ne i$, and vice versa. The loss for each sentence pair is calculated as follows: 

\begin{equation}
\begin{split}
\mathcal{L}_{alignment}(\theta) = \frac{1}{2B} \sum\limits_{i=1}^B (&log \frac{\exp(sim(s^{\scriptscriptstyle\parallel}_i, t^{\scriptscriptstyle\parallel}_i)/T)}{\sum\limits_{j=1}^{B} \exp(sim(s^{\scriptscriptstyle\parallel}_i, t^{\scriptscriptstyle\parallel}_j)/T)} \\ + &log \frac{\exp(sim(s^{\scriptscriptstyle\parallel}_i, t^{\scriptscriptstyle\parallel}_i)/T)}{\sum\limits_{j=1}^{B} \exp(sim(s^{\scriptscriptstyle\parallel}_j, t^{\scriptscriptstyle\parallel}_i)/T)}) \label{eq2}
\end{split}
\end{equation}

\noindent where $T$ is a temperature hyper-parameter, $sim(s,t)$ measures the cosine similarity between $s$ and $t$, and $B$ is the size of the aligned word pairs list. Note that some words may not be aligned in the word alignments derived from GIZA\texttt{++}, i.e., emerged from the insertion or deletion of tokens in the translation. These words are not considered in this loss function since it only iterates over the aligned word pairs. Thus, our fine-tuning has little effect on the representation of words that are not aligned to any other words. Moreover, by using high-precision word alignments taken from GIZA\texttt{++}, we try to guide the model only for confident aligned words. While \cref{eq2} is defined for a whole batch in \cite{wudredze2020}, we consider it for each sentence pair separately, as described later. Also, to prevent a degenerative solution, as in \cite{wudredze2020}, we use a regularization term, constraining parameters to stay close to their original pre-trained values,

\begin{equation}
\mathcal{L}_{reg}(\theta) = \left\| \theta - \theta_{pretrained} \right\|_2^2 \label{eq3} \end{equation}

Thus, the final objective will be as follows, with $\lambda = 1$ in the experiments:

\begin{equation}
\mathcal{L}(\theta) = \mathcal{L}_{alignment}(\theta) + \lambda\mathcal{L}_{reg}(\theta) \label{eq4}
\end{equation}

There are two aspects that our fine-tuning approach differs from \cite{wudredze2020}. First, they used the first subword of each word as its representative and did not consider other subwords in fine-tuning. However, we consider all the subwords of each aligned word pair as aligned to each other, as in our application, all subword representations are involved in computing XLMRScore. Thus, the model tries to represent all the subwords in the aligned word pairs closer to each other during fine-tuning. 

The second difference is negative examples, i.e., the unaligned tokens in the denominator of \cref{eq2}. \cite{wudredze2020} consider the unaligned tokens in each batch as negative examples. However, it is an improper choice as there might be two semantically similar words in two different sentence pairs in a batch, and their approach makes their embeddings further apart while fine-tuning the model, as they are not aligned to each other. For this reason, we consider the unaligned tokens in each sentence pair instead of each batch as negative examples, which is a better choice.

\cref{fig1}(a) shows an example of a mismatching error, where the Persian token ``\<بالا>'' which means ``up'', is matched to the English token ``The'' while using the XLMR-base model. However, after fine-tuning the model with our cross-lingual alignment strategy, it is matched to the correct word ``up'' as shown in \cref{fig1}(b).

\section{Datasets}\label{sec4}

In this section, we introduce the datasets that we used in this paper.
First, we present our newly English$\rightarrow$Persian test set with HTER labels as one of our main contributions and describe its preparation process.
Next, we introduce the WMT QE test sets that we used in our experiments to evaluate our proposed method.
Finally, we provide information about parallel and monolingual datasets that we utilized for fine-tuning the pre-trained model and building dictionaries for replacing untranslated tokens.

\subsection{English-Persian test data}\label{sec4.1}

Besides using the WMT21 QE test sets for evaluating our proposed method, we prepare and will publish the QE test set for English$\rightarrow$Persian (En-Fa) language pair on the HTER task. To the best of our knowledge, this is the first test set for the aforesaid language pair. Here we introduce this new test set and briefly describe its preparation process.

We use a collection of internal datasets, including scientific English papers in various domains, to obtain our QE test data. These papers were firstly translated from English to Persian using a commercial MT system named Faraazin\footnote{\url{https://www.faraazin.ir/}}. It was an RNN-based encoder-decoder model trained using the MarianNMT\footnote{\url{https://marian-nmt.github.io/}} toolkit \citep{junczys-dowmunt-etal-2018} and with a corpus of about 2.6 million sentence pairs. The vocabulary size was set to 100,000, the embedding size was 512, and the dimension of the hidden layer was 1024. The model was optimized using Adam optimizer, and the default MarianNMT values were used for the other parameters. The quality of this MT system in terms of BLEU score was 26.64 over a private test set with one reference from the AFEC corpora \citep{jabbari2012}. It could also get the BLEU score of 53.5\% over our prepared post-edited test set in this section. Note that the latter score is much larger, since unlike the former one, the references in it are post-edited from the MT outputs with minimum edits.

The MT outputs for 37 documents with a total of 3800 sentences were attained. Each document was given to a professional human translator to post-edit the MT output and provide the correct translations. Because certain post-edits were not of good quality, a second annotator checked all these source sentences, MT outputs, and human post-edits to select 1000 sentences with high-quality post-edits. Thus, for each sentence, we have one post-edit annotation, which is double-checked by another annotator. To calculate final QE HTER scores, we follow the guidelines of sentence-level post-editing effort task published in the WMT21 QE shared task~\citep{specia2021findings}. This task concerns scoring translations according to the proportion of the words that need to be edited to obtain a correct translation. The scores are generated by calculating the minimum edit distance between the MT output and its human post-edited version via HTER. The HTER labels are computed after tokenization using TERCOM\footnote{\url{https://github.com/jhclark/tercom}} \citep{snover2006}, with default settings.

\begin{table}[t]
\begin{center}
\begin{minipage}{280pt}
\caption{Statistics of the En-Fa test set}\label{tab1}
\begin{tabular}{@{}cccc@{}}
\toprule
\textbf{Sentences} & \textbf{Source Tokens} & \textbf{Target Tokens} & \textbf{Average HTER}\\
\midrule
1000    & 27,441   & 29,771  & 29.64  \\
\botrule
\end{tabular}
\end{minipage}
\end{center}
\end{table}

\begin{figure}[ht]%
\centering
\captionsetup{justification=raggedright,width=0.6\textwidth}
\includegraphics[width=0.6\textwidth]{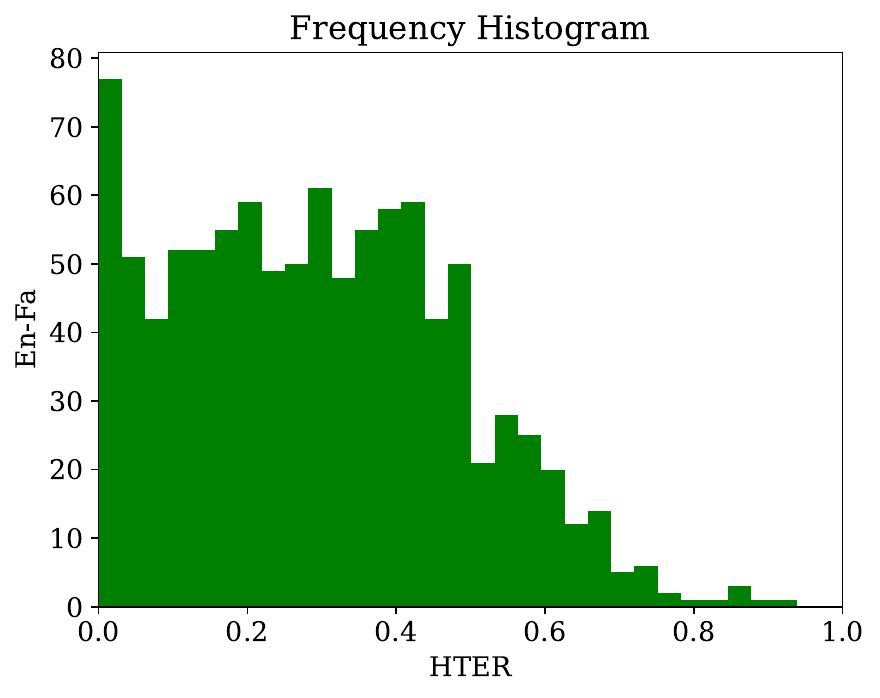}
\caption{Distribution of the HTER scores in the En-Fa test set.}
\label{fig2}
\end{figure}

\cref{tab1} shows the statistics of the provided test set, including the number of tokens on the source side and the MT output (after tokenization). The average MT quality, in terms of the HTER score, shows that the MT outputs have fairly good quality. The distribution of the sentence-level HTER scores is also demonstrated in \cref{fig2}. It shows that most translations are of reasonable quality with less than 0.4 HTER, while about 30\% of them get more than 0.4 HTER. This indicates that there are sufficient representatives of both high and low-quality translations in the test dataset.

We also analyze the distribution of different types of translation errors while computing the HTER scores. These errors include insertion, deletion, substitution, and shifts which are the edits required to change the MT outputs into the human post-edits. \cref{tab2} shows the distribution of these errors in our test corpus by indicating the amount and percentage of each error type over the whole corpus. It suggests that most of these errors were substitutions, i.e., wrongly translated words replaced by the correct ones in the post-edits.

\begin{table}[ht]
\begin{center}
\begin{minipage}{300pt}
\caption{Distribution of edit operations in the En-Fa test set, computed using the HTER metric}.\label{tab2}
\begin{tabular}{@{}c|cccc@{}}
\toprule
\textbf{Errors} & \textbf{Insertion} & \textbf{Deletion}  & \textbf{Substitution} & \textbf{Shift}\\
\midrule
\textbf{Number (Percentage)} & 1414 (15\%)    & 2112 (23\%)   & 4952 (53\%)  & 886 (9\%)  \\
\botrule
\end{tabular}
\end{minipage}
\end{center}
\end{table}

\begin{figure}[t]%
\centering
\captionsetup{justification=raggedright,width=0.6\textwidth}
\includegraphics[width=0.6\textwidth]{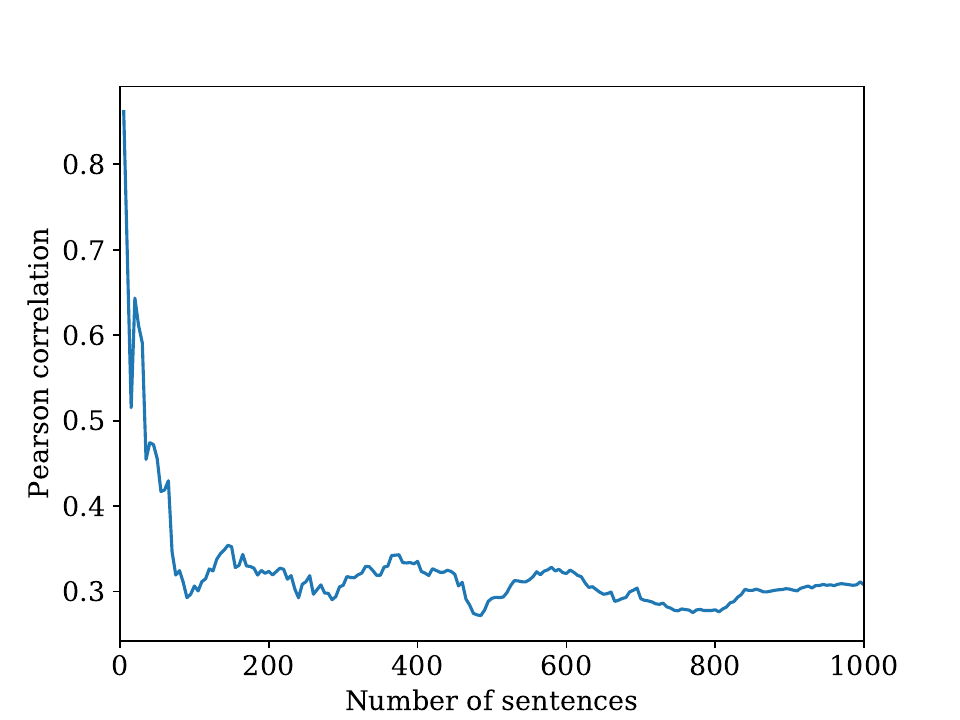}
\caption{Pearson correlation for our base model based on the number of sentences}
\label{fig3}
\end{figure}

To indicate that the size of test data is sufficient for evaluating a given QE method and adding more sentences to it will not affect the results significantly, we assess our base metric, i.e., the XLMRScore, on different sizes of test data in terms of the Pearson correlation with HTER labels. 
\cref{fig3} illustrates the results. As can be seen, the Pearson correlation changes drastically with a small number of sentences, while it is rather smooth and converges to a fixed number for test sizes above 850 sentences. Thus, we can conclude that 1000 sentences are sufficient for testing our model with reliable results, and the Pearson correlation would not dramatically change by adding more sentences. This size, i.e., 1000 sentences, is also analogous to the WMT test sets for the QE task.


\subsection{WMT data}\label{sec4.2}

For evaluating our unsupervised QE approach, we consider the following low resource language pairs from the WMT21 QE datasets\footnote{\url{https://github.com/sheffieldnlp/mlqe-pe}}: Nepalese$\rightarrow$English (Ne-En), Sinhala$\rightarrow$English (Si-En), Pashto$\rightarrow$English (Ps-En) and Khmer$\rightarrow$English (Km-En). The translation directions for these language pairs are all into English as indicated by the arrow. From these, only Ne-En and Si-En have training and development data, while Ps-En and Km-En were zero-shot scenarios with no training data available. We have only used the test data of all these language pairs for evaluations, as our approach is fully unsupervised. 

All these language pairs have annotated data for the Direct Assessment and HTER subtasks. We evaluate our method on both subtasks for these language pairs, as well as the new English$\rightarrow$Persian (En-Fa) test set provided and introduced in the previous section.

\subsection{Parallel and monolingual corpora}\label{sec4.3}

For the fine-tuning step in our proposed approach, we need a parallel corpus for each of the intended language pairs to get word alignments from statistical aligner tools. Thus, for Ne-En and Si-En, we use the parallel corpora provided by \cite{guzman2019flores}, and for Ps-En and Km-En, we use the clean parallel corpora provided by the WMT20 shared task on parallel corpus filtering\footnote{\url{https://www.statmt.org/wmt20/parallel-corpus-filtering.html}} \citep{koehn2020findings}. For the En-Fa language pair, we use the AFEC corpus \citep{jabbari2012}. In addition, we use the parallel IITB Hindi-English corpus\footnote{\url{https://www.cfilt.iitb.ac.in/iitb_parallel/}} \citep{kunchukuttan2018iit} for fine-tuning our cross-lingual model on Hindi as a related language to our intended languages. The statistics of the parallel corpora used can be found in \cref{tab3}.

\begin{table}[h]
\begin{center}
\begin{minipage}{195pt}
\caption{Statistics of the parallel corpora.}\label{tab3}%
\begin{tabular}{cccc}
\toprule
& \textbf{Languages} & \textbf{Sentences} & \textbf{Tokens} \\
\midrule
\multirow{6}{*}{\textbf{Parallel}} & Ne-En\footnotemark[1] & 495K & 2M \\
& Si-En & 647K & 3.7M \\
& Ps-En & 123K & 662K \\
& Km-En & 290K & 3.9M \\
& En-Fa\footnotemark[2] & 500K & 10.5M \\
& Hi-En\footnotemark[2] & 500K & 6.3M \\
\midrule
\end{tabular}
\footnotetext{The number of tokens for parallel corpora are reported over the English tokens.}
\footnotetext[1]{Only the GNOME/KDE/Ubuntu part was used.}
\footnotetext[2]{Only 500K sentences of the corpora were used, while the whole corpora is larger.}
\end{minipage}
\end{center}
\end{table}

For the strategy to replace the untranslated tokens proposed in \cref{subsec3.2}, we need monolingual corpora of the target language to build the required vocabulary. For this purpose, we use the English News crawl corpus\footnote{\url{https://data.statmt.org/news-crawl/}} for language pairs with English as the target side and the Persian side of the AFEC corpus \citep{jabbari2012} for the English$\rightarrow$Persian language pair. To build the target vocabulary, we consider the words with more than two occurrences in the monolingual corpora. The statistics of the monolingual corpora, in addition to the size of the created vocabulary, can be found in \cref{tab4}. As evident in \cref{tab4}, the vocabulary size highly depends on the size of the monolingual corpus used, and for English, it is much larger than Persian, as the monolingual corpus is much larger.

\begin{table}[h]
\begin{center}
\begin{minipage}{270pt}
\caption{Statistics of the monolingual corpora and vocabulary.}\label{tab4}%
\begin{tabular}{ccccc}
\toprule
& \textbf{Languages} & \textbf{Sentences} & \textbf{Tokens} & \textbf{Vocab size} \\
\midrule
\multirow{2}{*}{\textbf{Monolingual}} & En & 39.4M & 1B & 761K \\
& Fa & 684K & 15.4M & 60K\\
\botrule
\end{tabular}
\end{minipage}
\end{center}
\end{table}

\section{Experimental Setup}\label{sec5}

To conduct the experiments, we use the \nth{9} layer of the XLMR and the \nth{10} layer of the Aligned-XLMR (the model after fine-tuning on word alignments) to generate the contextual word representations since our experiments have shown that these layers could align words better in parallel sentences (\ref{sec7.1}). We use the Moses tokenizer \citep{koehn2007moses} for tokenizing the monolingual and parallel corpora while using GIZA\texttt{++} toolkit\footnote{\url{https://github.com/moses-smt/giza-pp}} to obtain word alignments. We evaluated the performance of our method in terms of Pearson correlation, using the WMT21 official evaluation scripts for sentence level tasks\footnote{\url{https://github.com/sheffieldnlp/qe-eval-scripts}}.

For fine-tuning the pre-trained model on word alignments, we use the implementation of \cite{wudredze2020} with our modifications on it. Moreover, we use the Adam optimizer \citep{kingma2015adam}, a learning rate of $10^{-4}$, batch size of 32, and the total steps of 100k for optimizing the model parameters during fine-tuning. The temperature hyperparameter ($T$) in the alignment loss function is set to $0.1$.



To analyze the performance of our proposed QE method, we have performed several experiments and present the results in this section. Each of the modifications to the baseline, i.e., replacing untranslated tokens and cross-lingual alignment, are examined separately to discuss its effectiveness. For the cross-lingual alignment strategy, we conduct our experiments both on a bilingual setting by fine-tuning the pre-trained model only on the word alignments of the intended source-target language pair (Aligned-XLMR-1-lang-pair) and on two multilingual settings by fine-tuning the model on the word alignments of all the 5 targeted language
pairs altogether (Aligned-XLMR-5-lang-pair), as well as adding Hindi-English to them (Aligned-XLMR-6-lang-pair). The proposed QE method and its modifications shown in the following tables are named as follows:

\begin{itemize}
\item \textbf{XLMR}: Scores obtained from the base XLMRScore. (\cref{subsec3.1})

\item \textbf{XLMR+Unk}: Scores obtained after replacing untranslated words with $\langle unk \rangle$. (\cref{subsec3.2})

\item \textbf{Aligned-XLMR-1-lang-pair}: Scores obtained after cross-lingual alignment of the XLMR-Base model only on the source-target language pair word alignments. (\cref{subsec3.3})

\item \textbf{Aligned-XLMR-1-lang-pair+Unk}: Scores obtained after both improvements of replacing untranslated tokens and cross-lingual alignment on the source-target language pair. (\cref{subsec3.2} and \cref{subsec3.3})

\item \textbf{Aligned-XLMR-5-lang-pair}: Scores obtained after cross-lingual alignment of the XLMR-Base model on all the 5 intended language pairs altogether. (\cref{subsec3.3})

\item \textbf{Aligned-XLMR-5-lang-pair+Unk}: Scores obtained after both improvements of replacing untranslated tokens and cross-lingual alignment on all the 5 intended language pairs. (\cref{subsec3.2} and \cref{subsec3.3})

\item \textbf{Aligned-XLMR-6-lang-pair}: Scores obtained after cross-lingual alignment of the XLMR-Base model on all the 5 intended language pairs as well as Hindi-English altogether. (\cref{subsec3.3})

\item \textbf{Aligned-XLMR-6-lang-pair+Unk}: Scores obtained after both improvements of replacing untranslated tokens and cross-lingual alignment on all the 5 intended language pairs as well as Hindi-English. (\cref{subsec3.2} and \cref{subsec3.3})
\end{itemize}

We conduct our experiments on two test sets: our En-Fa test set introduced in \cref{sec4.1} and the WMT21 QE Shared Task test sets for the aforementioned low-resource language pairs introduced in \cref{sec4.2}. In the following section, we present the results for each of these test sets separately.

\section{Results}\label{sec6}
\subsection{En-Fa Results}\label{sec6.1}

The resulting Pearson correlation for our proposed methods on the En-Fa test set is shown in \cref{tab5}. To validate the performance of our method, we compare our results with QE scores derived from two popular multilingual sentence embeddings, i.e., LASER\footnote{\url{https://github.com/facebookresearch/LASER}}  (Artetxe, 2018) and Sentence-BERT\footnote{\url{https://www.sbert.net/}}  (Nils, 2019, 2020), as the baseline models. The cosine similarity scores between the sentence embeddings of the source sentence and the hypothesis are considered as the QE scores for these models. 

The results show that both replacing untranslated words with the $\langle unk \rangle$ token (XLMR+Unk) and cross-lingual alignment of the pre-trained model (Aligned-XLMR-1-lang-pair) improve the Pearson correlation of our QE metric. Using both of these improvements together (Aligned-XLMR-1-lang-pair+Unk) enhances the correlation compared to the base XLMRScore by about 18\%. Also, \cref{tab5} shows that fine-tuning the pre-trained model on word alignments of the 5 intended language pairs altogether (Aligned-XLMR-5-lang-pair), as well as adding Hi-En to them (Aligned-XLMR-6-lang-pair), could further improve the results. This indicates that using the knowledge of related languages could help the model represent words in the targeted languages better. Furthermore, combining it with the strategy of replacing untranslated tokens in our final metric (Aligned-XLMR-6-lang-pair+Unk) gives the best results while outperforming the sentence-level embedding baselines by more than 50\%.

\begin{table}[h]
\begin{center}
\begin{minipage}{170pt}
\caption{Pearson correlation for the En-Fa test set.}\label{tab5}%
\begin{tabular}{p{130pt}|p{30pt}}
\toprule
\textbf{Method} & \textbf{En-Fa} \\
\hline \hline
\multicolumn{2}{c}{\parbox[c][2em]{\hsize{\textbf{Baselines}}}} \\
\midrule
LASER & 0.162 \\
Sentence-BERT & 0.24 \\
\midrule
\multicolumn{2}{c}{\textbf{Ours}} \\
\midrule
XLMR & 0.302 \\
XLMR+Unk & 0.338 \\
Aligned-XLMR-1-lang-pair & 0.315 \\
Aligned-XLMR-1-lang-pair+Unk & 0.356 \\
Aligned-XLMR-5-lang-pair & 0.329 \\
Aligned-XLMR-5-lang-pair+Unk & 0.358 \\
Aligned-XLMR-6-lang-pair & 0.34 \\ 
Aligned-XLMR-6-lang-pair+Unk & \textbf{0.369} \\
\botrule
\end{tabular}
\end{minipage}
\end{center}
\end{table}

\subsection{WMT21 Results}\label{sec6.2}

For the WMT21 test data, we evaluate our methods on both tasks of post-editing effort (HTER) and Direct Assessment to investigate how our proposed method performs for either of these tasks. Experimental results are shown in \cref{tab6}. Similar to the En-Fa results, the outcomes show that both strategies of replacing untranslated tokens and cross-lingual alignment of the pre-trained model would be beneficial and improve the Pearson correlation for almost all the language pairs and both HTER and DA tasks. It can also be seen that fine-tuning the model on word alignments from all the languages together (Aligned-XLMR-6-lang-pair) helps the method get better results for all the language pairs. Furthermore, incorporating it with the replacement strategy of untranslated tokens (Aligned-XLMR-6-lang-pair+Unk) gives the best results for almost all the language pairs, as for En-Fa in the previous section.

\begin{table}[t]
\begin{center}
\begin{minipage}{338pt}
\setlength{\tabcolsep}{1.52pt}
\caption{Pearson correlations with the HTER and Direct Assessment Tasks of WMT21 for Ps-En, Km-En, Ne-En, and Si-En language pairs}\label{tab6}%
\begin{tabular}{l|cccc|cccc}
\toprule
& \multicolumn{4}{c|}{\textbf{HTER}} & \multicolumn{4}{c}{\textbf{DA}} \\
& Ps-En & Km-En & Ne-En & Si-En & Ps-En & Km-En & Ne-En & Si-En \\
\hline \hline
\multicolumn{9}{c}{\parbox[c][2em]{\hsize{\textbf{Supervised}}}} \\
\midrule
WMT21 Baseline\footnotemark[1] & 0.503 & 0.576 & 0.626 & 0.607 & 0.476 & 0.562 & 0.738 & 0.513 \\
\midrule
\multicolumn{9}{c}{\textbf{Unsupervised}} \\
\midrule
SMOB-ECEIIT\footnotemark[1] & N/A & N/A & N/A & N/A & 0.424 & 0.409 & 0.544 & 0.347 \\
\midrule
\multicolumn{9}{c}{\textbf{Ours}} \\
\midrule
XLMR & 0.396 & 0.493 & 0.463 & 0.422 & 0.414 & 0.442 & 0.503 & 0.351 \\
XLMR+Unk & 0.401 & 0.494 & 0.460 & 0.410 & 0.422 & 0.448 & 0.498 & 0.354 \\
Aligned-XLMR-1-lang-pair & 0.413 & 0.558 & 0.526 & 0.455 & 0.444 & 0.486 & 0.512 & 0.376 \\
Aligned-XLMR-1-lang-pair+Unk & 0.418 & 0.560 & 0.523 & 0.431 & 0.454 & 0.494 & 0.514 & 0.381  \\
Aligned-XLMR-5-lang-pair & 0.424 & 0.565 & 0.528 & 0.468 & 0.452 & 0.484 & 0.540 & 0.382 \\
Aligned-XLMR-5-lang-pair+Unk & 0.428 & 0.567 & 0.518 & 0.443 & 0.463 & 0.491 & 0.532 & 0.384 \\
Aligned-XLMR-6-lang-pair & 0.425 & 0.575 & \textbf{0.542} & \textbf{0.475} & 0.457 & 0.489 & 0.550 & 0.386  \\
Aligned-XLMR-6-lang-pair+Unk & \textbf{0.432} & \textbf{0.577} & 0.540 & 0.455 & \textbf{0.467} & \textbf{0.496} & \textbf{0.551} & \textbf{0.397}  \\
\botrule
\end{tabular}
\footnotetext[1]{Results are taken from \cite{specia2021findings}}
\end{minipage}
\end{center}
\end{table}

To compare our results with other methods, \cref{tab6} also includes the results of the WMT21 supervised baseline, as well as the only fully unsupervised method that participated in WMT21, from \cite{specia2021findings} (i.e., SMOB-ECEIIT). The WMT21 baseline is a supervised transformer-based predictor-estimator model \citep{kim2017predictor, kepler2019openkiwi}, which uses the concatenated training portions of the WMT21 data (all seven language pairs together) for training on the corresponding task scores. Thus, it performs much better than our unsupervised method. However, we could get very close (less than 0.01 difference in Pearson correlation) to this supervised baseline in two of the zero-shot settings where no training data was available (i.e., Km-En in the HTER Task and Ps-En in the DA Task). Furthermore, our best results could surpass the WMT21 unsupervised participant for all the intended language pairs, although it was provided only for the DA Task. 

In SMOB-ECEIIT, they proposed two methods to compute distances between the source and translated candidate sentences and combined them linearly to get the final QE scores. Although no analysis of their outputs is available, the errors in the similarity of aligned words (the mismatching issue in BERTScore), in addition to the problem of untranslated words, can also negatively affect their method since they used the XLMR-base model to compute distances between words. Thus, not only do our final results outperform the SMOB-ECEIIT, but our two proposed strategies of cross-lingual alignment of the pre-trained model and replacing the untranslated tokens can also benefit their method.

\section{Discussion}\label{sec7}

In this section, we further analyze the effects of fine-tuning the pre-trained model on related languages and the limitations of our replacement strategy for untranslated tokens. We also discuss the results of our unsupervised approach in comparison to the SOTA supervised methods, as well as our results in the WMT22 Explainable QE Task.

\subsection{Effects of multilingual fine-tuning of the pre-trained model}\label{sec7.1}

One of the experiments that we have done is fine-tuning the pre-trained model in multilingual settings on the word alignments of all the targeted language pairs, as well as Hindi-English. We expect that as these languages are close and somehow related to each other, like Pashto and Persian or Nepalese and Hindi, fine-tuning the pre-trained model on them simultaneously could help the model use the knowledge of related languages. Thus, we fine-tune the XLMR-Base model to optimize the cross-lingual alignment loss defined in \cref{eq4} on all these language pairs together. The multilingual fine-tuning results are also shown in \cref{tab5} and \cref{tab6}, named Aligned-XLMR-5-lang-pair and Aligned-XLMR-6-lang-pair. As expected, the Aligned-XLMR-6-lang-pair model, which is fine-tuned on all the 6 language pairs, including Hi-En, gives the best results while improving the previous outcomes for all language pairs.

In order to study the effects of our cross-lingual alignment described in \cref{subsec3.3} more directly, we also evaluate our best fine-tuned model in the word alignment task, using the greedy matching method described in \cite{sabet2020simalign}. We conduct evaluations with golden alignments of En-Fa \citep{tavakoli2014} and En-Hi\footnote{\url{http://web.eecs.umich.edu/~mihalcea/wpt05/}} language pairs and use the \textit{Alignment Error Rate} (AER) \citep{ochney2003} as the evaluation metric.

\begin{figure}[t]%
\centering

\medskip

\begin{subfigure}[t]{.49\linewidth}
\centering\includegraphics[width=\linewidth]{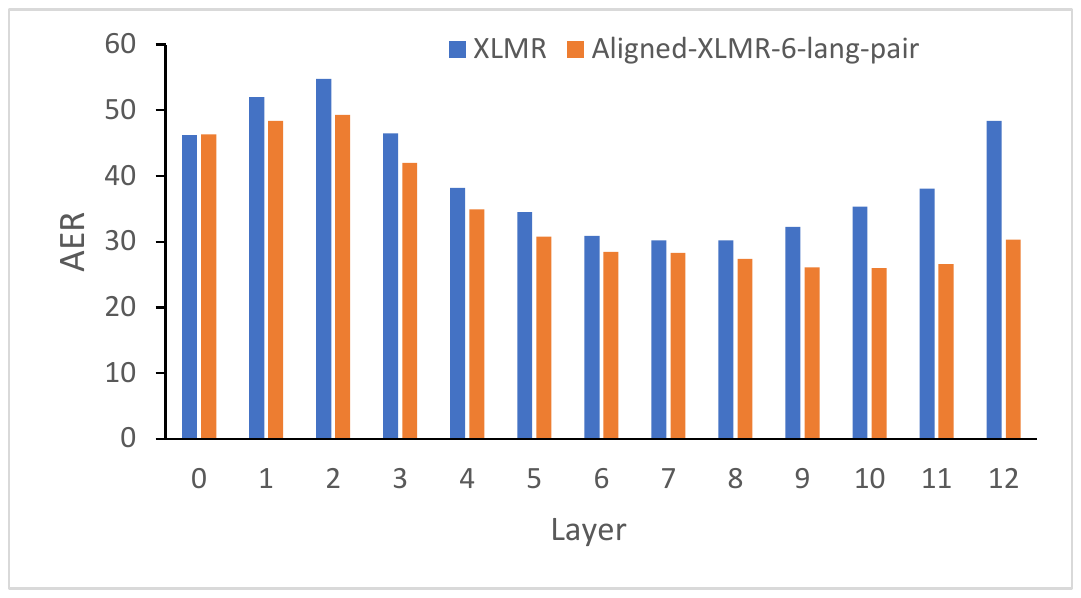}
\caption{En-Fa}
\end{subfigure}
\begin{subfigure}[t]{.49\linewidth}
\centering\includegraphics[width=\linewidth]{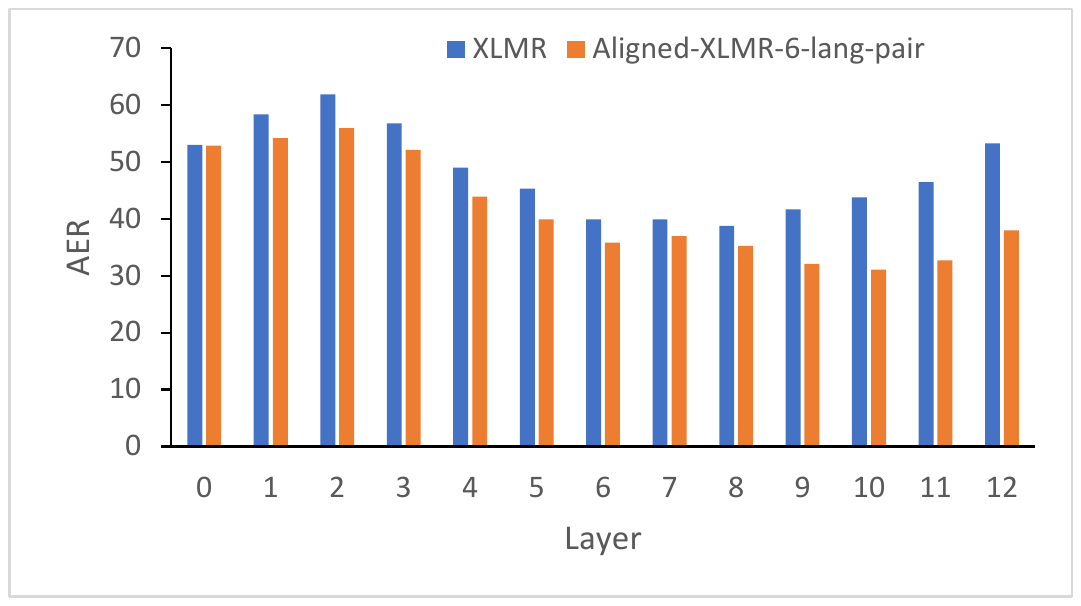}
\caption{En-Hi}
\end{subfigure}
\caption{The AER Results across different layers for En-Fa (a) and En-Hi (b) test sets}
\label{fig4}
\end{figure}

Figure 5 illustrates the AER while using the representations of each layer in the word alignment task for the XLMR-Base model and its cross-lingually aligned version after fine-tuning on all the 6 language pairs. It shows that for both language pairs, the AER significantly reduces in all the layers after fine-tuning, specifically for the last few layers. This shows that the cross-lingually aligned model improves the alignment of words in parallel sentences. As a result, it enhances matched words in the greedy matching step while computing XLMRScore, and thus it could alleviate the mismatching issue as expected.

\subsection{Limitations of replacing untranslated tokens}\label{sec7.2}
As described in \cref{subsec3.2}, we use a vocabulary of the target language derived from a monolingual corpus to detect the untranslated tokens in the translation candidate. A limitation of this approach is that the comprehensiveness of this vocabulary and the effectiveness of this strategy strongly depend on the monolingual corpora used. If the monolingual corpus is large enough, we expect the vocabulary to contain common named entity translations, perfect cognates, or abbreviations that do not need to be translated from source to target language.

Although using this strategy has shown to be beneficial in our experiments, our investigations reveal that the created vocabulary is not so comprehensive, and there are some tokens in the test sets, including the named entities, abbreviations, or even some rather uncommon words in the target language that are not in the vocabulary and are replaced and tagged as unknown tokens wrongly, which could mislead the QE method. Thus, future attempts to improve the identification of untranslated tokens, such as by enhancing the target language vocabulary or proposing a language identification mechanism to the pre-trained model, could be promising.

\subsection{Comparing to SOTA supervised methods}\label{sec7.3}

Although our unsupervised approach to QE is still inferior to the SOTA supervised methods that participated in the WMT21 QE shared task, it is still relevant as an attempt to move ahead the unsupervised approaches in this field, which has been rather under-explored to date. According to \cite{specia2021findings}, using model ensembling and glass-box features from the NMT model are two strategies occurring in most of the SOTA QE systems, such as QEMind \citep{wang2021} and IST-Unbabel \citep{zerva-etal-2021-ist}. Although we did not target these strategies in this paper, adding them to our unsupervised method can close its performance gap with the SOTA supervised methods in the future.

Also, as indicated in \cite{specia-etal-2020} and \cite{specia2021findings}, a noteworthy fact about the performance of supervised methods on low-resource and zero-shot language pairs is that they rely heavily on the target language, i.e., English in all of our intended language pairs in the WMT21 datasets. Thus, they benefit from the shared target language between the available training data and the low-resource language pairs test data. In other words, these models could learn to estimate the quality of English sentences from the training data they used, i.e., concerning fluency, grammatical correctness, etc., and this might be the reason for their high performance even on zero-shot language pairs with no training data available. Accordingly, their performance degrades for low-resource non-English target languages, and thus unsupervised approaches can be beneficial for these scenarios. However, confirming this is left for future work, as none of the WMT21 test sets in our experiments have non-English target languages.

\subsection{Explainable QE}\label{sec7.4}
One advantage of our unsupervised approach is that its quality scores are easily explainable; that is, we can easily understand which erroneous words in the translation candidate lead to such bad sentence-level quality scores. This is due to the fact that in XLMRScore, we simply average over the cosine similarities of the matched target and source tokens, and thus a target token with a low similarity score with its matched source token is more likely to be an error. This motivated us to participate in the WMT22 shared task on Explainable QE \citep{zerva-etal-2022}, and in this section, we present our results for this task.

This task aims to address translation error identification as rationale extraction from sentence-level quality estimation systems \citep{zerva-etal-2022}. Specifically, the participating teams are asked to provide word-level scores in addition to the sentence-level scores for each sentence pair so that the tokens with the highest scores are expected to correspond to translation errors. It is different from the word-level QE task as it is not allowed to supervise the models with any token or word-level labels or signals, and the scores must be extracted as an explanation from the sentence-level QE system. Finally, these token-level scores are evaluated and compared to the human word-level error annotations using the Recall at Top-k metric.

\begin{table}[t]
\begin{center}
\begin{minipage}{200pt}
\setlength{\tabcolsep}{3pt}
\caption{Recall at Top-K for the WMT22 Explainable QE Task for Ps-En and Km-En language pairs. Results are taken from \cite{zerva-etal-2022}}\label{tab7}%
\begin{tabular}{l|cc}
\toprule
\textbf{Model} & \textbf{Ps-En} & \textbf{Km-En} \\
\hline \hline
BASELINE (Random)  & 0.614 & 0.565 \\
BASELINE (OpenKiwi+LIME) & 0.615 & 0.580 \\
IST-Unbabel & 0.672 & 0.665 \\
HW-TSC & \textbf{0.715} & \textbf{0.686}   \\
\midrule
Our Method & 0.668 & 0.622 \\
\botrule
\end{tabular}
\end{minipage}
\end{center}
\end{table}

For this task, we participate in Km-En and Ps-En language pairs and simply report the negative of the cosine similarity of each target token to its matched source token as its quality score. The results in \cref{tab7} indicated that our approach outperforms the WMT22 baselines, including the baseline with the supervised OpenKiwi. This shows that although our unsupervised approach could not achieve the performance of SOTA supervised models in sentence-level QE, due to its inherent explainability, it could get close to the supervised methods in this task and even surpass the OpenKiwi with the LIME post-hoc explanation tool baseline. Furthermore, the interesting results of the cosine similarities as token quality scores can motivate proposing other unsupervised methods which use them to attain better sentence-level QE scores. 

\section{Conclusion and Future Work}\label{sec8}

In this work, we have presented an unsupervised approach for translation QE. The main idea of our method, called XLMRScore, is based on using the well-known BERTScore metric, which was originally proposed for the monolingual evaluation tasks, in the cross-lingual scenario of QE. For this purpose, we use a multilingual pre-trained model, namely XLMR-Base, in our method while indicating and tackling the issues of using this metric directly for QE. We have proposed two strategies to alleviate these issues, i.e., replacing out-of-vocabularies (OOVs) with the $\langle unk \rangle$ token for the untranslated words issue and cross-lingual alignment of the pre-trained model for the so-called mismatching issue. We have conducted our experiments on 4 low-resource languages of the WMT21 test sets, which show that both proposed strategies significantly improve the base QE model, with average improvements of above 8\% for both HTER and DA tasks. Remarkably, our final model could get comparable results to the supervised baseline in two zero-shot scenarios while surpassing the existing unsupervised method by above 11\% on average. In addition, we have introduced a new English$\rightarrow$Persian QE test set with the post-editing effort task labels (HTER), which could be used for zero-shot QE evaluations by other researchers, and we have tested our proposed method on it as well.

We consider five avenues for future work. First, we will further investigate other fine-tuning or pre-training strategies to have a better cross-lingually aligned pre-trained model. Second, we plan to use our cross-lingually aligned model as the predictor in the SOTA supervised methods to understand if it is beneficial for supervised scenarios as well. Third, adding a mechanism for language identification to pre-trained models instead of replacing untranslated words could also be advantageous to study. Fourth, it is also worth adding the recurring strategies in the SOTA supervised models, such as using glass-box features and model ensembling, to our unsupervised approach, as they can significantly increase its performance. And fifth, the performance of unsupervised approaches should be compared with supervised ones for test sets with non-English target languages to confirm their effectiveness for such scenarios.


\bibliography{sn-bibliography}

\begin{thebibliography}{51}
\providecommand{\natexlab}[1]{#1}
\providecommand{\url}[1]{{#1}}
\providecommand{\urlprefix}{URL }
\providecommand{\doi}[1]{\url{https://doi.org/#1}}
\providecommand{\eprint}[2][]{\url{#2}}
 \bibcommenthead

\bibitem[{Aldarmaki and Diab(2019)}]{aldarmakidiab2019}
Aldarmaki H, Diab M (2019) Context-aware cross-lingual mapping. In: Proceedings of the 2019 Conference of the North {A}merican Chapter of the Association for Computational Linguistics: Human Language Technologies, Volume 1 (Long and Short Papers). Association for Computational Linguistics, Minneapolis, Minnesota, pp 3906--3911

\bibitem[{Banerjee and Lavie(2005)}]{banerjee2005}
Banerjee S, Lavie A (2005) Meteor: An automatic metric for mt evaluation with improved correlation with human judgments. In: Proceedings of the ACL workshop on intrinsic and extrinsic evaluation measures for machine translation and/or summarization, pp 65--72

\bibitem[{Cao et~al(2019)Cao, Kitaev, and Klein}]{cao2019}
Cao S, Kitaev N, Klein D (2019) Multilingual alignment of contextual word representations. In: International Conference on Learning Representations

\bibitem[{do~Carmo et~al(2021)do~Carmo, Shterionov, Moorkens, Wagner, Hossari, Paquin, Schmidtke, Groves, and Way}]{do2021review}
do~Carmo F, Shterionov D, Moorkens J, et~al (2021) A review of the state-of-the-art in automatic post-editing. Machine Translation 35(2):101--143

\bibitem[{Chen et~al(2021)Chen, Su, Zhang, Wang, Geng, Yang, Tao, Jiaxin, Minghan, Zhang et~al}]{chen2021}
Chen Y, Su C, Zhang Y, et~al (2021) Hw-tsc’s participation at wmt 2021 quality estimation shared task. In: Proceedings of the Sixth Conference on Machine Translation, pp 890--896

\bibitem[{Conneau and Lample(2019)}]{conneau2019cross}
Conneau A, Lample G (2019) Cross-lingual language model pretraining. Advances in neural information processing systems 32

\bibitem[{Conneau et~al(2020)Conneau, Khandelwal, Goyal, Chaudhary, Wenzek, Guzm{\'a}n, Grave, Ott, Zettlemoyer, and Stoyanov}]{conneau2020unsupervised}
Conneau A, Khandelwal K, Goyal N, et~al (2020) Unsupervised cross-lingual representation learning at scale. In: Proceedings of the 58th Annual Meeting of the Association for Computational Linguistics. Association for Computational Linguistics, Online, pp 8440--8451

\bibitem[{Cuturi(2013)}]{cuturi2013}
Cuturi M (2013) Sinkhorn distances: Lightspeed computation of optimal transport. Advances in neural information processing systems 26

\bibitem[{Devlin et~al(2019)Devlin, Chang, Lee, and Toutanova}]{devlin2019bert}
Devlin J, Chang MW, Lee K, et~al (2019) {BERT}: Pre-training of deep bidirectional transformers for language understanding. In: Proceedings of the 2019 Conference of the North {A}merican Chapter of the Association for Computational Linguistics: Human Language Technologies, Volume 1 (Long and Short Papers). Association for Computational Linguistics, Minneapolis, Minnesota, pp 4171--4186

\bibitem[{Edelsbrunner and Morozov(2012)}]{Edelsbrunner2012}
Edelsbrunner H, Morozov D (2012) Persistent homology: Theory and practice. In: Proceedings of the European Congress of Mathematics. European Mathematical Society, pp 31--50

\bibitem[{Etchegoyhen et~al(2018)Etchegoyhen, Garcia, and Azpeitia}]{etchegoyhen2018}
Etchegoyhen T, Garcia EM, Azpeitia A (2018) Supervised and unsupervised minimalist quality estimators: Vicomtech’s participation in the wmt 2018 quality estimation task. In: Proceedings of the Third Conference on Machine Translation: Shared Task Papers, pp 782--787

\bibitem[{Fomicheva et~al(2020)Fomicheva, Sun, Yankovskaya, Blain, Guzm{\'a}n, Fishel, Aletras, Chaudhary, and Specia}]{fomicheva2020unsupervised}
Fomicheva M, Sun S, Yankovskaya L, et~al (2020) Unsupervised quality estimation for neural machine translation. Transactions of the Association for Computational Linguistics 8:539--555

\bibitem[{Fomicheva et~al(2022)Fomicheva, Sun, Fonseca, Zerva, Blain, Chaudhary, Guzm{\'a}n, Lopatina, Specia, and Martins}]{fomicheva2022mlqe}
Fomicheva M, Sun S, Fonseca E, et~al (2022) {MLQE}-{PE}: A multilingual quality estimation and post-editing dataset. In: Proceedings of the Thirteenth Language Resources and Evaluation Conference. European Language Resources Association, Marseille, France, pp 4963--4974

\bibitem[{Guzm{\'a}n et~al(2019)Guzm{\'a}n, Chen, Ott, Pino, Lample, Koehn, Chaudhary, and Ranzato}]{guzman2019flores}
Guzm{\'a}n F, Chen PJ, Ott M, et~al (2019) The {FLORES} evaluation datasets for low-resource machine translation: {N}epali{--}{E}nglish and {S}inhala{--}{E}nglish. In: Proceedings of the 2019 Conference on Empirical Methods in Natural Language Processing and the 9th International Joint Conference on Natural Language Processing (EMNLP-IJCNLP). Association for Computational Linguistics, Hong Kong, China, pp 6098--6111

\bibitem[{Huang et~al(2019)Huang, Liang, Duan, Gong, Shou, Jiang, and Zhou}]{huang2019unicoder}
Huang H, Liang Y, Duan N, et~al (2019) {U}nicoder: A universal language encoder by pre-training with multiple cross-lingual tasks. In: Proceedings of the 2019 Conference on Empirical Methods in Natural Language Processing and the 9th International Joint Conference on Natural Language Processing (EMNLP-IJCNLP). Association for Computational Linguistics, Hong Kong, China, pp 2485--2494

\bibitem[{Ive et~al(2018)Ive, Blain, and Specia}]{ive2018}
Ive J, Blain F, Specia L (2018) Deepquest: a framework for neural-based quality estimation. In: Proceedings of the 27th International Conference on Computational Linguistics, pp 3146--3157

\bibitem[{Jabbari et~al(2012)Jabbari, Bakshaei, Ziabary, and Khadivi}]{jabbari2012}
Jabbari F, Bakshaei S, Ziabary SMM, et~al (2012) Developing an open-domain english-farsi translation system using afec: Amirkabir bilingual farsi-english corpus. In: Fourth Workshop on Computational Approaches to Arabic-Script-based Languages, pp 17--23

\bibitem[{Junczys-Dowmunt et~al(2018)Junczys-Dowmunt, Grundkiewicz, Dwojak, Hoang, Heafield, Neckermann, Seide, Germann, Aji, Bogoychev, Martins, and Birch}]{junczys-dowmunt-etal-2018}
Junczys-Dowmunt M, Grundkiewicz R, Dwojak T, et~al (2018) {M}arian: Fast neural machine translation in {C}++. In: Proceedings of {ACL} 2018, System Demonstrations. Association for Computational Linguistics, Melbourne, Australia, pp 116--121

\bibitem[{K et~al(2020)K, Wang, Mayhew, and Roth}]{k-etal-2020}
K K, Wang Z, Mayhew S, et~al (2020) Cross-lingual ability of multilingual {BERT:} an empirical study. In: 8th International Conference on Learning Representations

\bibitem[{Kepler et~al(2019{\natexlab{a}})Kepler, Tr{\'e}nous, Treviso, Vera, G{\'o}is, Farajian, Lopes, and Martins}]{kepler-etal-2019-unbabels}
Kepler F, Tr{\'e}nous J, Treviso M, et~al (2019{\natexlab{a}}) Unbabel{'}s participation in the {WMT}19 translation quality estimation shared task. In: Proceedings of the Fourth Conference on Machine Translation (Volume 3: Shared Task Papers, Day 2). Association for Computational Linguistics, Florence, Italy, pp 78--84

\bibitem[{Kepler et~al(2019{\natexlab{b}})Kepler, Tr{\'e}nous, Treviso, Vera, and Martins}]{kepler2019openkiwi}
Kepler F, Tr{\'e}nous J, Treviso M, et~al (2019{\natexlab{b}}) Openkiwi: An open source framework for quality estimation. In: Proceedings of the 57th Annual Meeting of the Association for Computational Linguistics: System Demonstrations, pp 117--122

\bibitem[{Kim et~al(2017)Kim, Jung, Kwon, Lee, and Na}]{kim2017predictor}
Kim H, Jung HY, Kwon H, et~al (2017) Predictor-estimator: neural quality estimation based on target word prediction for machine translation. ACM Transactions on Asian and Low-Resource Language Information Processing (TALLIP) 17(1):1--22

\bibitem[{Kim et~al(2019)Kim, Lim, Kim, and Na}]{kim2019}
Kim H, Lim JH, Kim HK, et~al (2019) Qe bert: bilingual bert using multi-task learning for neural quality estimation. In: Proceedings of the Fourth Conference on Machine Translation (Volume 3: Shared Task Papers, Day 2), pp 85--89

\bibitem[{Kingma and Ba(2015)}]{kingma2015adam}
Kingma DP, Ba J (2015) Adam: {A} method for stochastic optimization. In: 3rd International Conference on Learning Representations, {ICLR}

\bibitem[{Koehn et~al(2007)Koehn, Hoang, Birch, Callison-Burch, Federico, Bertoldi, Cowan, Shen, Moran, Zens, Dyer, Bojar, Constantin, and Herbst}]{koehn2007moses}
Koehn P, Hoang H, Birch A, et~al (2007) {M}oses: Open source toolkit for statistical machine translation. In: Proceedings of the 45th Annual Meeting of the Association for Computational Linguistics Companion Volume Proceedings of the Demo and Poster Sessions. Association for Computational Linguistics, Prague, Czech Republic, pp 177--180

\bibitem[{Koehn et~al(2020)Koehn, Chaudhary, El-Kishky, Goyal, Chen, and Guzm{\'a}n}]{koehn2020findings}
Koehn P, Chaudhary V, El-Kishky A, et~al (2020) Findings of the {WMT} 2020 shared task on parallel corpus filtering and alignment. In: Proceedings of the Fifth Conference on Machine Translation. Association for Computational Linguistics, Online, pp 726--742

\bibitem[{Kulshreshtha et~al(2020)Kulshreshtha, Redondo~Garcia, and Chang}]{kulshreshtha-etal-2020}
Kulshreshtha S, Redondo~Garcia JL, Chang CY (2020) Cross-lingual alignment methods for multilingual {BERT}: A comparative study. In: Findings of the Association for Computational Linguistics: EMNLP 2020. Association for Computational Linguistics, pp 933--942

\bibitem[{Kunchukuttan et~al(2018)Kunchukuttan, Mehta, and Bhattacharyya}]{kunchukuttan2018iit}
Kunchukuttan A, Mehta P, Bhattacharyya P (2018) The {IIT} {B}ombay {E}nglish-{H}indi parallel corpus. In: Proceedings of the Eleventh International Conference on Language Resources and Evaluation ({LREC} 2018). European Language Resources Association (ELRA), Miyazaki, Japan

\bibitem[{Lee(2020)}]{lee2020}
Lee D (2020) Two-phase cross-lingual language model fine-tuning for machine translation quality estimation. In: Proceedings of the Fifth Conference on Machine Translation, pp 1024--1028

\bibitem[{Liu et~al(2019)Liu, McCarthy, Vuli{\'c}, and Korhonen}]{liu2019}
Liu Q, McCarthy D, Vuli{\'c} I, et~al (2019) Investigating cross-lingual alignment methods for contextualized embeddings with token-level evaluation. In: Proceedings of the 23rd Conference on Computational Natural Language Learning (CoNLL). Association for Computational Linguistics, Hong Kong, China, pp 33--43

\bibitem[{Moura et~al(2020)Moura, Vera, van Stigt, Kepler, and Martins}]{moura2020}
Moura J, Vera M, van Stigt D, et~al (2020) Ist-unbabel participation in the wmt20 quality estimation shared task. In: Proceedings of the Fifth Conference on Machine Translation, pp 1029--1036

\bibitem[{Och and Ney(2003)}]{ochney2003}
Och FJ, Ney H (2003) A systematic comparison of various statistical alignment models. Computational Linguistics 29(1):19--51

\bibitem[{Papineni et~al(2002)Papineni, Roukos, Ward, and Zhu}]{papineni2002}
Papineni K, Roukos S, Ward T, et~al (2002) Bleu: a method for automatic evaluation of machine translation. In: Proceedings of the 40th annual meeting of the Association for Computational Linguistics, pp 311--318

\bibitem[{Ranasinghe et~al(2020)Ranasinghe, Orǎsan, and Mitkov}]{ranasinghe2020}
Ranasinghe T, Orǎsan C, Mitkov R (2020) Transquest: Translation quality estimation with cross-lingual transformers. In: Proceedings of the 28th International Conference on Computational Linguistics, pp 5070--5081

\bibitem[{Sabet et~al(2020)Sabet, Dufter, Yvon, and Sch{\"u}tze}]{sabet2020simalign}
Sabet MJ, Dufter P, Yvon F, et~al (2020) Simalign: High quality word alignments without parallel training data using static and contextualized embeddings. In: Findings of the Association for Computational Linguistics: EMNLP 2020, pp 1627--1643

\bibitem[{Snover et~al(2006)Snover, Dorr, Schwartz, Micciulla, and Makhoul}]{snover2006}
Snover M, Dorr B, Schwartz R, et~al (2006) A study of translation edit rate with targeted human annotation. In: Proceedings of the 7th Conference of the Association for Machine Translation in the Americas: Technical Papers, pp 223--231

\bibitem[{Specia et~al(2009)Specia, Turchi, Cancedda, Cristianini, and Dymetman}]{specia2009}
Specia L, Turchi M, Cancedda N, et~al (2009) Estimating the sentence-level quality of machine translation systems. In: Proceedings of the 13th annual conference of the European association for machine translation

\bibitem[{Specia et~al(2013)Specia, Shah, De~Souza, and Cohn}]{specia2013}
Specia L, Shah K, De~Souza JG, et~al (2013) Quest-a translation quality estimation framework. In: Proceedings of the 51st Annual Meeting of the Association for Computational Linguistics: System Demonstrations, pp 79--84

\bibitem[{Specia et~al(2020)Specia, Blain, Fomicheva, Fonseca, Chaudhary, Guzm{\'a}n, and Martins}]{specia-etal-2020}
Specia L, Blain F, Fomicheva M, et~al (2020) Findings of the {WMT} 2020 shared task on quality estimation. In: Proceedings of the Fifth Conference on Machine Translation. Association for Computational Linguistics, Online, pp 743--764

\bibitem[{Specia et~al(2021)Specia, Blain, Fomicheva, Zerva, Li, Chaudhary, and Martins}]{specia2021findings}
Specia L, Blain F, Fomicheva M, et~al (2021) Findings of the {WMT} 2021 shared task on quality estimation. In: Proceedings of the Sixth Conference on Machine Translation. Association for Computational Linguistics, pp 684--725

\bibitem[{Tavakoli and Faili(2014)}]{tavakoli2014}
Tavakoli L, Faili H (2014) Phrase alignments in parallel corpus using bootstrapping approach. International Journal of Information and Communication Technology Research

\bibitem[{Tuan et~al(2021)Tuan, El-Kishky, Renduchintala, Chaudhary, Guzm{\'a}n, and Specia}]{tuan2021}
Tuan YL, El-Kishky A, Renduchintala A, et~al (2021) Quality estimation without human-labeled data. In: Proceedings of the 16th Conference of the European Chapter of the Association for Computational Linguistics: Main Volume, pp 619--625

\bibitem[{Wang et~al(2018)Wang, Fan, Li, Zhou, Chen, Shi, and Si}]{wang2018}
Wang J, Fan K, Li B, et~al (2018) Alibaba submission for wmt18 quality estimation task. In: Proceedings of the Third Conference on Machine Translation: Shared Task Papers, pp 809--815

\bibitem[{Wang et~al(2021)Wang, Wang, Chen, Zhao, Luo, and Zhang}]{wang2021}
Wang J, Wang K, Chen B, et~al (2021) Qemind: Alibaba’s submission to the wmt21 quality estimation shared task. In: Proceedings of the Sixth Conference on Machine Translation. Association for Computational Linguistics, pp 948--954

\bibitem[{Wang et~al(2019)Wang, Che, Guo, Liu, and Liu}]{wang2019}
Wang Y, Che W, Guo J, et~al (2019) Cross-lingual {BERT} transformation for zero-shot dependency parsing. In: Proceedings of the 2019 Conference on Empirical Methods in Natural Language Processing and the 9th International Joint Conference on Natural Language Processing (EMNLP-IJCNLP). Association for Computational Linguistics, Hong Kong, China, pp 5721--5727

\bibitem[{Wu and Dredze(2019)}]{wu-dredze-2019-beto}
Wu S, Dredze M (2019) Beto, bentz, becas: The surprising cross-lingual effectiveness of {BERT}. In: Proceedings of the 2019 Conference on Empirical Methods in Natural Language Processing and the 9th International Joint Conference on Natural Language Processing (EMNLP-IJCNLP). Association for Computational Linguistics, Hong Kong, China, pp 833--844

\bibitem[{Wu and Dredze(2020)}]{wudredze2020}
Wu S, Dredze M (2020) Do explicit alignments robustly improve multilingual encoders? In: Proceedings of the 2020 Conference on Empirical Methods in Natural Language Processing (EMNLP). Association for Computational Linguistics, Online, pp 4471--4482

\bibitem[{Zerva et~al(2021)Zerva, van Stigt, Rei, Farinha, Ramos, C.~de Souza, Glushkova, Vera, Kepler, and Martins}]{zerva-etal-2021-ist}
Zerva C, van Stigt D, Rei R, et~al (2021) {IST}-unbabel 2021 submission for the quality estimation shared task. In: Proceedings of the Sixth Conference on Machine Translation. Association for Computational Linguistics, Online, pp 961--972

\bibitem[{Zerva et~al(2022)Zerva, Blain, Rei, Lertvittayakumjorn, C.~de Souza, Eger, Kanojia, Alves, Or{\u{a}}san, Fomicheva, Martins, and Specia}]{zerva-etal-2022}
Zerva C, Blain F, Rei R, et~al (2022) Findings of the {WMT} 2022 shared task on quality estimation. In: Proceedings of the Seventh Conference on Machine Translation (WMT). Association for Computational Linguistics, pp 69--99

\bibitem[{Zhang et~al(2020)Zhang, Kishore, Wu, Weinberger, and Artzi}]{zhang2020bertscore}
Zhang T, Kishore V, Wu F, et~al (2020) Bertscore: Evaluating text generation with bert. In: International Conference on Learning Representations

\bibitem[{Zhou et~al(2020)Zhou, Ding, and Takeda}]{zhou2020zero}
Zhou L, Ding L, Takeda K (2020) Zero-shot translation quality estimation with explicit cross-lingual patterns. In: Proceedings of the Fifth Conference on Machine Translation. Association for Computational Linguistics, Online, pp 1068--1074

\end{thebibliography}


\end{document}